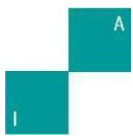

# INTELIGENCIA ARTIFICIAL

# Exploring Machine Learning and Transformer-based Approaches for Deceptive Text Classification: A Comparative Analysis

Anusuya Krishnan[1]

[1] College of Information Technology, United Arab Emirates University, UAE

**Abstract** Deceptive text classification is a critical task in natural language processing that aims to identify deceptive or fraudulent content. This study presents a comparative analysis of machine learning and transformer-based approaches for deceptive text classification. We investigate the effectiveness of traditional machine learning algorithms and state-of-the-art transformer models, such as BERT, XLNET, DistilBERT, and RoBERTa, in detecting deceptive text. A labeled dataset consisting of deceptive and non-deceptive texts is used for training and evaluation purposes. Through extensive experimentation, we compare the performance metrics, including accuracy, precision, recall, and F1 score, of the different approaches. The results of this study shed light on the strengths and limitations of machine learning and transformer-based methods for deceptive text classification, enabling researchers and practitioners to make informed decisions when dealing with deceptive content.

**Resumen** La clasificación de texto engañoso es una tarea crítica en el procesamiento del lenguaje natural que tiene como objetivo identificar contenido engañoso o fraudulento. Este estudio presenta un análisis comparativo del aprendizaje automático y los enfoques basados en transformadores para la clasificación engañosa de textos. Investigamos la efectividad de los algoritmos tradicionales de aprendizaje automático y los modelos de transformadores de última generación, como BERT, XLNET, DistilBERT y RoBERTa, para detectar texto engañoso. Se utiliza un conjunto de datos etiquetados que consta de textos engañosos y no engañosos con fines de capacitación y evaluación. A través de una amplia experimentación, comparamos las métricas de rendimiento, incluida la exactitud, la precisión, la recuperación y la puntuación F1, de los diferentes enfoques. Los resultados de este estudio arrojan luz sobre las fortalezas y limitaciones del aprendizaje automático y los métodos basados en transformadores para la clasificación engañosa de textos, lo que permite a los investigadores y profesionales tomar decisiones informadas cuando se trata de contenido engañoso.

## 1 Introduction

The continuous expansion of e-commerce platforms has fueled a significant increase in the exchange of opinions and the abundance of online reviews regarding products. Leading platforms like FlipKart, Amazon, Talabat, Careem, and Zoom empower individuals to share their opinions and provide valuable insights for potential buyers. When considering online purchases, customers heavily rely on these reviews to gain a deeper understanding of the products they wish to purchase. By analyzing the opinions of fellow customers on e-commerce websites, new buyers can make well-informed decisions about whether to proceed with a purchase. Positive feedback from online reviews serves as a powerful motivator, influencing customers to confidently invest in a product and underscoring the importance of these reviews as an essential source of information. Unfortunately, amidst the authenticity of genuine reviews, deceptive practices have emerged, wherein individuals intentionally post misleading reviews to either promote or undermine the reputation of specific products. These misleading reviews, commonly known as fake reviews, have the potential to distort customer perceptions and manipulate their purchasing decisions [1].

Furthermore, businesses themselves may engage in this practice by generating inauthentic content to manipulate customers' opinions. Individuals who partake in posting deceptive opinions are often referred to as opinion spammers. However, despite the growth in the number of online reviews, the prevalence of fake reviews has outpaced the overall improvement in their quality. This escalation of malicious false reviews has resulted in a rising number of instances where both retailers and customers suffer harm. Consequently, users are increasingly faced with the challenge of distinguishing helpful reviews from the overwhelming flood of information. The blurring of the intrinsic value of online reviews, which traditionally aids in reducing uncertainty during pre-purchase decisions, has led to a decline in the credibility and traffic of e-commerce platforms [1].

Online reviews play a crucial role in providing customers with valuable information about products they intend to buy. Customers willingly share their experiences, both positive and negative, which can significantly impact businesses in the long term. However, this environment also opens the door for the manipulation of customer decisions through the generation of false or fake reviews, commonly known as opinion spamming. Spammers purposefully craft deceptive opinions to sway others, with the intention of either boosting or damaging the reputation of a business or product [2]. Deceptive reviews can be categorized into three main groups. The first group includes reviews that contain intentionally false information about a product, aiming to manipulate its reputation. The second group consists of reviews that focus on the brand itself, without expressing any actual experience with a specific product. The third group encompasses non-reviews and advertisements that contain text indirectly related to the product. It is relatively straightforward to identify the second and third groups, but detecting the first group can be more challenging. These reviews may be authored by individual spammers hired by business owners or a collective effort of spammers working together within a specific timeframe to manipulate the reputation of a product or store [3-4].

Furthermore, it has become evident that opinion spamming extends beyond product reviews and customer feedback. This paper emphasizes the importance of online reviews as critical sources of information for customers in the e-commerce industry. It also highlights the challenges posed by deceptive reviews, which aim to mislead and manipulate consumers. Understanding the impact of fake reviews and the presence of opinion spammers is vital for preserving the integrity and credibility of online reviews, ensuring that customers can make well-informed purchasing decisions [4]. Deceptive text classification refers to the task of identifying and distinguishing deceptive or fraudulent content from genuine text. Deception in text can manifest in various forms, such as fake reviews, fraudulent advertisements, or misleading information. Detecting and accurately classifying deceptive text is crucial for ensuring the credibility and reliability of textual information in various domains, including online platforms, social media, and customer feedback systems. The rise of digital platforms and the ease of online communication have provided opportunities for individuals and organizations to manipulate or deceive others through textual content. Traditional methods of deceptive text classification typically rely on manual inspection or rule-based approaches, which can be time-consuming, labor-intensive, and limited in their ability to handle large volumes of text.

In recent years, the emergence of machine learning and natural language processing techniques has opened new avenues for more effective and efficient deceptive text classification. Machine learning algorithms can automatically learn patterns and features from labeled datasets, enabling them to identify subtle linguistic cues that indicate deception. Furthermore, transformer-based models, such as XLNET (EXtreme Learning Network.), DistilBERT (Distilled Bidirectional Encoder Representations from Transformers), BERT (Bidirectional Encoder Representations from Transformers) and RoBERTa (Robustly Optimized Bidirectional Encoder Representations from Transformers Approach), have demonstrated exceptional performance in various natural language processing tasks, including sentiment analysis and text classification. These models have the potential to capture the contextual information and semantic relationships necessary for accurately detecting deceptive text. This brief introduction sets the stage for further exploration and investigation into deceptive text classification techniques. By leveraging the power of machine learning and transformer-based approaches, researchers and practitioners can develop more robust and efficient methods for identifying deceptive content, thereby contributing to the maintenance of trust, transparency, and integrity in digital communication and information dissemination.

This paper is structured as follows to provide a comprehensive overview of the research findings. Section 2 offers a summary of related works in the field, drawing insights from existing literature. In Section 3, we provide a detailed exploration of the background and intricacies of our proposed machine learning approach for detecting deceptive reviews. Moving on to Section 4, we present the specifics of two experiments conducted to evaluate the accuracy of our model in identifying deceptive reviews. The outcomes of these experiments are thoroughly discussed, offering valuable insights into the effectiveness of our approach. Lastly, in Section 5, we conclude the paper by summarizing the key findings and contributions of our work. Additionally, we highlight potential avenues for future research and development.

## 2 Related works

In this segment, we explore existing works on deceptive text detection and the diverse methodologies employed by researchers to identify deceptive text. The deceptive text dataset, which served as the pioneer dataset for deceptive reviews in Amazon reviews, established a benchmark for their detection [6-7]. Research in deceptive text detection can be classified into two main approaches: review spam (textual) and review spammer (behavioral). One study concentrated on the textual-based detection of deceptive reviews. The authors employed n-gram analysis and term frequency as feature extraction techniques to identify deceptive reviews in a dataset collected from Amazon Mechanical Turk and TripAdvisor. The dataset encompassed deceptive reviews of Chicago hotels, alongside honest reviews categorized into positive and negative groups. Their approach achieved an accuracy of 86% using a support vector machine (SVM) classifier [1-4].

One study focused on analyzing reviews with a high number of reviewer's votes and comments to identify suspicious and potentially fake reviews. The researchers hypothesized that reviews with fewer votes are more likely to be fake. They employed supervised machine learning techniques such as SVM, NB, and LR to detect review spam. The NB method yielded the highest F-score of 0.58, outperforming other methods that relied on behavioral features [10]. Other studies explored the correlation between distribution anomalies and the detection of fake reviews. They suggested that certain businesses may hire spammers to generate fake reviews. To evaluate this, they utilized an opinion spam "gold-standard dataset" consisting of 400 deceptive and truthful reviews. Their approach achieved an accuracy of 72.5% on the test dataset, demonstrating the effectiveness of detecting suspicious bursts within a specific time window. However, their method was less effective in determining the authenticity of user reviews [12].

Another study developed a deceptive model to identify suspicious spammers based on their behavioral patterns. They categorized reviewers as spammers or non-spammers and proposed an unsupervised bayesian inference framework for detecting deceptive reviews. The results showed that the deceptive model outperformed other supervised machine learning approaches, highlighting its efficiency in identifying deceptive reviews [13]. The study suggested the potential use of K-Nearest Neighbors (KNN) algorithms to improve the model's accuracy. However, limitations were noted regarding the versatility of the emotional dictionary, which was specialized in the movie domain. Additionally, the study did not incorporate adverbs, which play a crucial role in representing the intensity of emotional expression words [14].

Extracting opinion data involves analyzing text data that expresses user opinions about an object or its specific features. In the case of Korean, this process involves preprocessing steps such as morphology analysis and phrase analysis. Through natural language processing techniques, including morpheme analysis and parsing, the opinion text is analyzed to extract information such as the expression of the opinion, the object being referred to, and any modifiers. The sentiment of specific words is determined as positive or negative, and the strength of the sentiment is adjusted to derive a final polarity value [15]. RLSA (Review based Linear Sentiment Analysis tool) considers the polarity information of opinion data as a factor comparable to a "measurement," such as sales volume. By utilizing the polarity information in the opinion data, RLSA captures complex information, including overall user evaluations of products, regional assessments, and temporal changes in opinions [15-16]. These studies offer diverse approaches to detect deceptive reviews, including burst detection mechanisms, analysis of singleton reviews, consideration of reviewer's votes, examination of distribution anomalies, and author spamicity modeling. Each approach provides valuable insights and contributes to the ongoing research in the field of deceptive review detection.

In another study, researchers expanded upon the work on the opinion spam dataset and utilized their own dataset to detect deceptive reviews [8]. However, they noted that machine-generated reviews do not accurately represent real-world deceptive reviews, as they lack the characteristic elements of opinion spam. When tested on a Yelp dataset, the model trained using machine-generated fake reviews only achieved a 67.8% accuracy, indicating its limitations in detecting real-world deceptive reviews. Nonetheless, the authors emphasized the continued value of n-gram features in identifying deceptive reviews. They also introduced a novel technique known as the burst detection mechanism, which aimed to identify customers who provide deceptive reviews [9]. This approach utilized a Markov random field model and implemented a loopy belief propagation method, resulting in a 77.6% accuracy.

Another approach focused on detecting singleton reviews (SR), as the study revealed that over 90% of reviewers only post a single review. Singleton reviews tend to have a significantly larger size compared to non-singleton reviews. The researchers also observed that a rapid increase in the number of singleton reviews within a short duration indicates potential manipulation by fake reviewers aiming to influence the reputation or rating of a product. In early research, the focus was on identifying deceptive opinions in online reviews. The authors introduced a method that combines linguistic and behavioral features to detect deceptive reviews. They conducted experiments using machine learning techniques, including SVM and Naive Bayes, and achieved promising results. Another study explored the use of recurrent neural networks, particularly LSTM networks, for detecting deceptive opinion spam. The authors incorporated syntactic, semantic, and stylistic features into the LSTM model, demonstrating its effectiveness in capturing deceptive patterns in online reviews [17].

In a separate investigation, researchers addressed the issue of fake review detection by employing machine learning algorithms. Their proposed approach combined content-based, stylistic, and behavioral features to identify deceptive reviews, showcasing the efficacy of their method through experimental results. In an effort to detect fake reviews without relying on labeled training data, researchers proposed an unsupervised learning approach. This method utilized review rating patterns and review text similarity to identify potential fake reviews, and it showcased promising results in detecting deceptive opinions [16].

Furthermore, researchers explored the application of transformer models, specifically BERT, for deceptive review detection. They fine-tuned the pre-trained BERT model on a dataset consisting of deceptive and genuine reviews, achieving state-of-the-art performance in deceptive text classification tasks. Another study aimed to tackle the challenge of detecting deceptive opinion spam across different domains. The authors proposed a domain-independent approach that leveraged content-based and behavior-based features, demonstrating the effectiveness of their method through experiments conducted in various domains. Although not solely focused on deceptive text in reviews, another study concentrated on detecting fake news using deep learning techniques. The authors employed a neural network model that captured both lexical and temporal features to identify fake news articles. Their approach yielded promising results in detecting deceptive textual information.

## 3 Proposed Methodology

The process of identifying fake reviews involves multiple steps. Initially, the dataset is preprocessed to remove unnecessary special characters, punctuation, stop words, and irrelevant terms. Subsequently, lemmatization is applied to extract meaningful features from the refined dataset. Finally, the classification process includes training a classifier using the extracted features. In our first study, we evaluated the performance of five distinct machine learning algorithms: support vector machine (SVM), linear support vector machines (LSVM), passive aggressive classifier (PA), logistic regression (LR), and multinomial naive Bayes (NB). In the second step, we extended our analysis by employing transformer models for feature extraction. Specifically, we utilized XLNET (EXtreme Learning Network), DistilBERT (Distilled Bidirectional Encoder Representations from Transformers), BERT (Bidirectional Encoder Representations from Transformers), and RoBERTa (Robustly Optimized Bidirectional Encoder Representations from Transformers Approach) to process the extracted features.

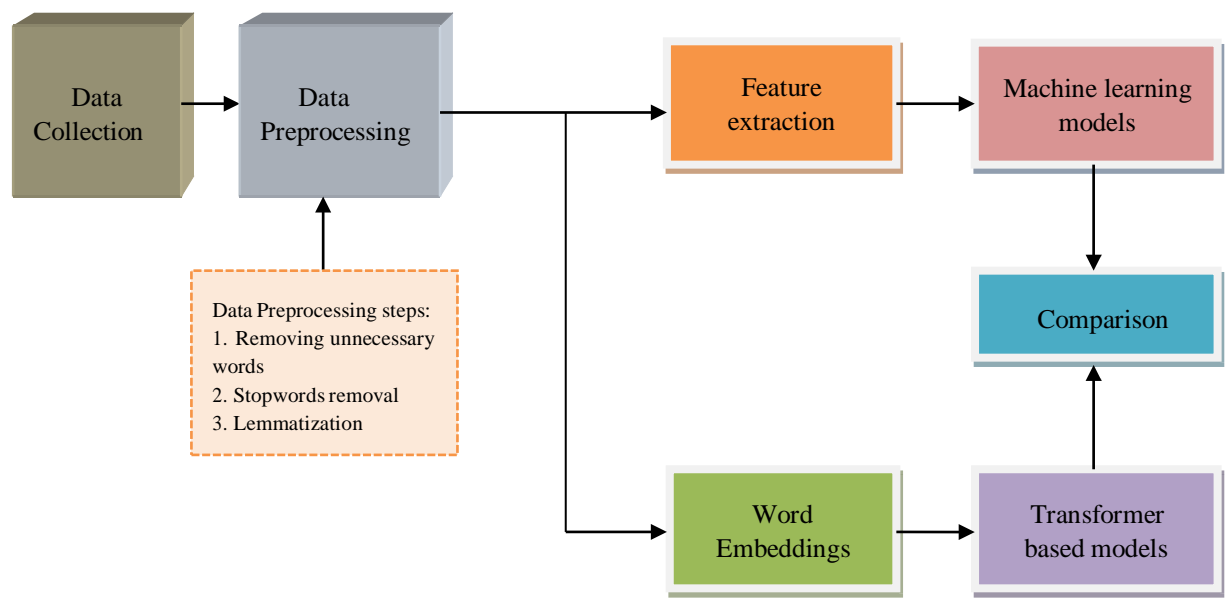

Figure 1. Proposed Methodology

### 3.1  Data Collection

In our study, we employed the deceptive detection dataset, which is a publicly available dataset sourced from Kaggle. This dataset comprises a total of 1600 reviews and has been previously utilized in research study [2]. The dataset is divided into two categories: 800 truthful reviews and 800 fake reviews. The focus of this dataset is specifically on the top twenty hotels in Chicago. The reviews within the dataset were collected from platforms such as TripAdvisor and Amazon Mechanical Turk. Each entry in the dataset provides various information, including the review label indicating its authenticity (whether it is fake or truthful), the corresponding hotel name, the sentiment or polarity of the review categorized as positive or negative, the review source (TripAdvisor or Mechanical Turk), and finally, the actual text of the review itself. This comprehensive dataset enables us to analyze and detect deceptive opinions effectively.

### 3.2  Data Preprocessing

In machine learning tasks, the significance of data preprocessing cannot be overstated, particularly when working with unstructured data. The primary objective of data preprocessing is to clean and refine the data using a range of techniques. These techniques encompass various operations such as removing punctuation, URLs, and stop words, as well as converting text to lowercase, tokenization, stemming, and lemmatization. By applying these techniques, irrelevant information is effectively eliminated, and the data is prepared for feature extraction. Tokenization, a fundamental technique in natural language processing, plays a crucial role in breaking down the text into smaller units called tokens. These tokens can include alphanumeric characters, punctuation marks, or special characters. For example, the sentence "the fruit is delicious" would be tokenized into individual tokens such as "the," "fruit," "is," and "delicious." Stop words, which include commonly used words like articles, conjunctions, prepositions, and pronouns, are typically devoid of significant meaning. Therefore, during the preprocessing stage, it is common practice to remove these stop words. This step helps streamline the data by eliminating unnecessary words that do not contribute substantially to the overall understanding of the text. Lemmatization is a vital process in transforming tokenized words into their base or root forms, thereby enhancing human comprehension. This technique reduces words to their common root form, disregarding variations in tense or form. For example, words like "studying," "studied," and "study" would all be reduced to their base form, which is "study." In our study, we recognize the value of lemmatization as an integral part of our data preprocessing approach. By employing these techniques, we ensure that the data is appropriately refined and prepared for subsequent stages of analysis. Through effective data preprocessing, we can enhance the quality of the data and facilitate more accurate and meaningful feature extraction, leading to improved machine learning outcomes.

### 3.3  Feature extraction

To input text data into our machine learning model, it is crucial to convert words into numerical or vector representations, requiring the use of feature extraction. In this study, we focused on a specific feature extraction method: term frequency-inverse document frequency (TF-IDF). While the bag of words approach is straightforward and effective, it treats all words equally without considering their relative importance. To overcome this limitation, we employed TF-IDF, a widely adopted feature extraction technique in natural language processing. TF-IDF evaluates the significance of a word within a document in relation to the entire corpus.

TF-IDF transforms words into vector form by multiplying the term frequency (TF) with the inverse document frequency (IDF). Term frequency (TF) is calculated by dividing the number of occurrences of a word in a document by the total number of words in that document. Inverse document frequency (IDF) is determined by taking the logarithm of the ratio between the total number of documents in the corpus and the number of documents in the corpus containing the specific word. The resulting TF-IDF score is obtained by multiplying the TF with the IDF. By utilizing TF-IDF, we address the limitations of the bag-of-words approach and effectively capture the importance of words within a document relative to the entire corpus. This enables us to represent the text data in a numerical format suitable for our machine learning model's utilization.

### 3.4  Word Embeddings

Word embeddings in transformer models, such as XLNET (EXtreme Learning Network), DistilBERT (Distilled Bidirectional Encoder Representations from Transformers), BERT (Bidirectional Encoder Representations from Transformers), and RoBERTa (Robustly Optimized Bidirectional Encoder Representations from Transformers Approach), play a critical role in capturing the meaning and context of words in natural language processing tasks. These transformer-based models leverage large-scale pre-training tasks to learn powerful contextualized word embeddings. Unlike traditional word embeddings, which represent words as fixed-size vectors, contextual word embeddings adapt to the context in which a word appears. They capture the influence of the surrounding words and the overall sentence or document structure, resulting in more nuanced and informative representations.

In transformer models, the embedding layer is responsible for mapping input words or subword units (such as subword tokens in BERT) to continuous vector representations. The embedding layer initializes word embeddings randomly at the beginning of training and updates them during the model's learning process. The transformer architecture incorporates attention mechanisms, allowing the model to attend to different parts of the input sequence during processing. Attention helps capture the relationships between words and enables the model to understand the context in which a word is used. When a transformer model is pre-trained on a large corpus, it learns contextualized word embeddings by predicting missing words in masked language modeling tasks or by generating coherent and meaningful text in autoregressive language modeling tasks. These pre-trained models capture rich semantic and syntactic information, making them highly effective in downstream natural language processing tasks. During fine-tuning, transformer models are further trained on task-specific data. The contextual word embeddings, which have learned representations from the pre-training stage, provide a strong foundation for understanding the semantics, nuances, and contextual information in the task-specific data. By utilizing contextual word embeddings in transformer models, these models excel in various natural language processing tasks, including language understanding, sentiment analysis, named entity recognition, text classification, machine translation, and question answering. The contextualized representations capture the intricate relationships between words, enabling the models to handle complex language patterns and improve the overall performance on a wide range of language-based tasks.

### 3.5   Machine model training

During the training phase of a machine learning model, it acquires knowledge from a labeled dataset to enhance its ability to make precise predictions or perform specific tasks. The training process involves iteratively providing the model with extracted feature data along with corresponding labels. This allows the model to adjust its internal parameters and optimize its performance over time. To facilitate the training process, we split the data into two components: features (input variables) and labels (output variables). This separation allows the model to learn the relationships between the input features and the corresponding output labels. We employ the "train_test_split" function to further divide the data into training and testing sets. This division ensures that we have a dedicated portion of the data to evaluate the model's performance after training.

We initialize machine learning models, namely support vector machine (SVM), linear support vector machines (LSVM), passive aggressive classifier (PA), logistic regression (LR), and multinomial naive Bayes (NB). These models possess different characteristics and capabilities, offering diverse approaches to tackle the given task. We train the selected model on the training data using the `fit` method. This process involves adjusting the model's internal parameters based on the patterns and relationships present in the training data. To assess the trained model's performance, we evaluate its accuracy using the `score` method on the testing data. The accuracy metric provides a measure of the model's predictive capability, indicating how well it generalizes to unseen data. Finally, we print the obtained accuracy score to gain insights into the model's effectiveness.

### 3.6   Transformer based model training

During the training phase of a Transformer-based model, we harness the power of self-attention mechanisms and large-scale pre-training to develop a robust and versatile language model. Transformer models, such as XLNET (EXtreme Learning Network), DistilBERT (Distilled Bidirectional Encoder Representations from Transformers), BERT (Bidirectional Encoder Representations from Transformers), and RoBERTa (Robustly Optimized Bidirectional Encoder Representations from Transformers Approach), have revolutionized natural language processing tasks. The training process typically involves two main steps: pre-training and fine-tuning. In the pre-training step, the model is trained on a massive corpus of unlabeled text. This unsupervised pre-training enables the model to learn the statistical properties of the language, capture syntactic and semantic relationships, and develop a contextual understanding of words.

During pre-training, the model is trained on tasks like masked language modeling (MLM) or autoregressive language modeling (e.g., predicting the next word in a sequence). By predicting missing or masked words, the model learns to encode contextual information and generate coherent representations for various language patterns. After pre-training, the model undergoes fine-tuning on specific downstream tasks. This involves training the model on labeled task-specific data. The labeled data could include sentiment analysis, named entity recognition, text classification, or question answering, among others. During fine-tuning, the model adapts its pre-trained knowledge to the task at hand, refining its internal parameters and learning task-specific patterns. The training data is typically split into training and validation sets. The model is trained on the training set, and its performance is evaluated on the validation set to monitor progress and prevent overfitting. Various optimization techniques, such as stochastic gradient descent (SGD) and Adam optimization, are employed to update the model's parameters iteratively and minimize the loss function. Throughout the training process, hyperparameter tuning is often performed to find the optimal settings for the model architecture, learning rate, batch size, and regularization. This fine-tuning allows the model to generalize well, improve its accuracy, and perform effectively on unseen data. Training a Transformer-based model requires substantial computational resources and time, as well as access to large-scale text data for pre-training. However, the result is a powerful language model that can understand context, handle complex language patterns, and excel in a wide range of natural language processing tasks.

### 3.7 Model Evaluation

Model evaluation metrics are essential for assessing the performance and effectiveness of a machine learning model. These metrics provide quantitative measures that help in understanding how well the model performs on specific tasks and enable comparisons between different models. When evaluating our machine learning model, we considered a range of evaluation metrics to gain a comprehensive understanding of its performance. The metrics used include accuracy, precision, recall, F1 score, and area under the receiver operating characteristic curve (AUC-ROC). Accuracy is a fundamental metric that measures the overall correctness of the model's predictions. It quantifies the proportion of correctly classified instances compared to the total number of instances. Precision is a metric that evaluates the model's ability to correctly identify positive instances among the predicted positives. It calculates the ratio of true positives to the sum of true positives and false positives. Recall, also known as sensitivity or true positive rate, measures the model's ability to correctly identify positive instances among all actual positive instances. It computes the ratio of true positives to the sum of true positives and false negatives. The F1 score is a harmonic mean of precision and recall. It provides a balanced measure that considers both precision and recall simultaneously. F1 score is particularly useful when there is an imbalance between the classes. The AUC-ROC is a metric commonly used for binary classification problems. It measures the model's ability to rank instances correctly by calculating the area under the receiver operating characteristic curve. A higher AUC-ROC indicates better model performance. By considering these evaluation metrics, we gain insights into different aspects of the model's performance. This comprehensive evaluation allows us to make informed decisions about the model's suitability for the given task. It provides a quantitative assessment of its strengths and weaknesses, helping us refine and improve the model for optimal performance in real-world scenarios.

## 4 Experimental results and discussion

In this section, we conducted an evaluation of our proposed approach using the deceptive text dataset [1] and subsequently discussed the results obtained.

### 4.1 Machine learning experimental results

For our study, we utilized the publicly available deceptive text on hotel reviews dataset, which can be accessed on Kaggle. This dataset comprises 1600 reviews that have been categorized into 800 truthful reviews and 800 fake reviews [2]. The reviews specifically focus on the top twenty hotels in Chicago. We collected the dataset from two sources: TripAdvisor and Amazon Mechanical Turk. During our analysis, we primarily focused on two key attributes from the dataset: the review label indicating its authenticity and the review text itself. Other attributes were disregarded during the data pre-processing phase to streamline the analysis. To transform the textual data into numerical representations, we employed the TF-IDF vectorizer, which assigns weights to words based on their frequency in the reviews. During the feature extraction process, we explored different configurations for the n-gram range and the maximum number of features. We discovered that the choice of n-gram range and the maximum number of features significantly influenced the accuracy of the models. Specifically, we observed that increasing the value of the n-gram range had a noticeable impact on the overall accuracy of the models. Additionally, we conducted experiments with various values for the maximum number of features, ranging from 1000 to 25000. Through evaluation and analysis, we determined that setting the n-gram range to (2,2) and the maximum number of features to 1000 yielded the best accuracy compared to other feature settings. After pre-processing and vectorizing the data, we proceeded to feed the transformed data into our machine learning models for further analysis and prediction.

The evaluation results for different classifiers using the TF-IDF feature extraction method are as follows: Logistic Regression (LR) achieved an accuracy of 87.8%, an F1 score of 87.6%, a recall of 87.7%, a precision of 87.8%, and an AUC of 91.04%. Linear Support Vector Machines (LSVM) achieved an accuracy of 90.1%, an F1 score of 90.0%, a recall of 90.4%, a precision of 89.9%, and an AUC of 96.34%. Passive Aggressive Classifier (PA) achieved an accuracy of 90.5%, an F1 score of 91.0%, a recall of 90.5%, a precision of 90.5%, and an AUC of 97.54%. Multinomial Naive Bayes (NB) achieved an accuracy of 86.7%, an F1 score of 87.0%, a recall of 85.7%, a precision of 87.6%, and an AUC of 94.56%. Support Vector Machines (SVM) achieved an accuracy of 89.5%, an F1 score of 89.5%, a recall of 89.0%, a precision of 89.5%, and an AUC of 95.98%. Table 1 explains the performance of five machine learning models using TF-IDF vectorizer.

Table.1. Performance of five machine learning models using TF-IDF vectorizer.

| Classifiers | Feature | Accuracy | F1 score | Recall | Precision | AUC |
| --- | --- | --- | --- | --- | --- | --- |
| LR | TF-IDF | 87.8% | 87.6% | 87.7% | 87.8% | 91.04% |
| LSVM | TF-IDF | 90.1% | 90% | 90.4% | 89.9% | 96.34% |
| PA | TF-IDF | 90.5% | 91% | 90.5% | 90.5% | 97.54% |
| NB | TF-IDF | 86.7% | 87% | 85.7% | 87.6% | 94.56% |
| SVM | TF-IDF | 89.5% | 89.5% | 89% | 89.5% | 95.98% |

These evaluation metrics provide valuable insights into the performance of each classifier when using the TF-IDF feature representation. The Passive Aggressive Classifier (PA) achieved the highest accuracy, precision, recall, and AUC scores, indicating its effectiveness in accurately detecting the target classification task. The Linear Support Vector Machines (LSVM) and Support Vector Machines (SVM) also demonstrated strong performance across multiple metrics. The Multinomial Naive Bayes (NB) classifier achieved slightly lower accuracy but still exhibited considerable performance. These results highlight the suitability of these classifiers for the given context and emphasize their capabilities in effectively distinguishing between classes based on the TF-IDF features.

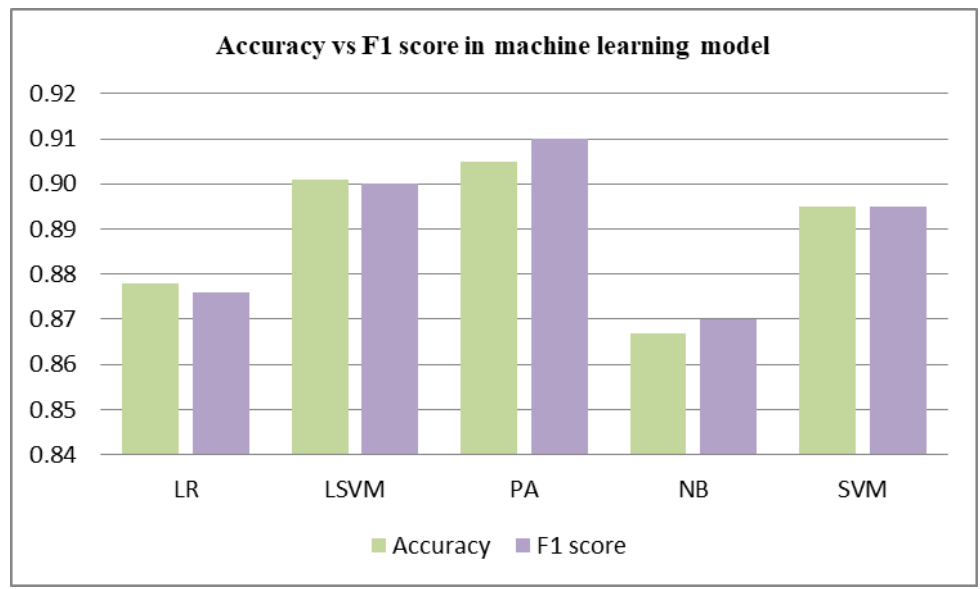

Figure 2. Comparison of all five machine learning model (accuracy vs F1 score) using barchart

Figure 2 presents a comprehensive overview of the classifier performance using the TF-IDF vectorizer in a train-test split. The figure clearly illustrates that the passive aggressive classifier outperforms all other classifiers, achieving an impressive accuracy of 90.5%. Additionally, the passive aggressive classifier consistently demonstrates strong performance across different train-test splits when utilizing the TF-IDF vectorizer. Figure 3 shows the chart of AUC- ROC curve vs precision vs recall. The graph shows that the PA model has the highest

AUC score, followed by the LR model, the LSVM model, and the NB model. The PA model also has the highest precision, followed by the SVM model, the LSVM model, and the LR model. The NB model has the lowest precision and recall. This suggests that the PA model is the best performing model overall, followed by the LSVM model. The LR model and the NB model are not as good performing models.

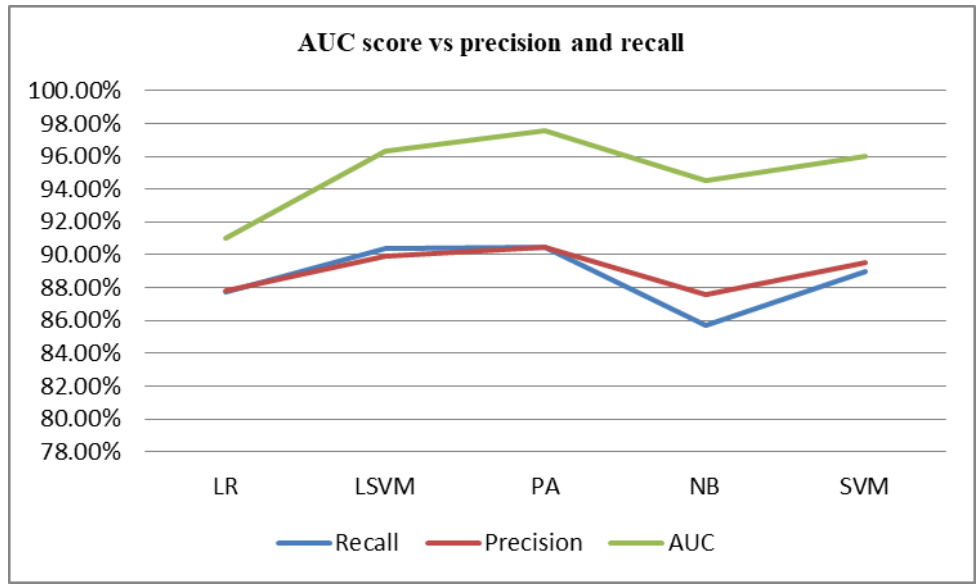

Figure 3. Comparison of all five machine learning model (AUC-ROC vs recall vs precision) using line graph

## 4.2 Transformer based results

After the preprocessing step, where the text data is cleaned and transformed into a suitable format, the next stage involves converting the words into word embeddings and utilizing them in a transformer model. Word embeddings are numerical representations of words that capture semantic relationships and meaning. They map words to dense vectors in a high-dimensional space, where words with similar meanings are closer together. These embeddings are obtained using techniques like Word2Vec, which train on large text corpora to learn the vector representations. In the context of a transformer model, such as BERT, XLNET, DistilBERT, or RoBERTa, the word embeddings are used as input. The transformer model further refines these embeddings by considering the context and interdependencies between words. It leverages attention mechanisms to capture the relationships between different words in a sequence, enabling the model to understand the contextual meaning of each word.

During training, the transformer model learns to predict missing or masked words based on the surrounding context, enabling it to capture intricate linguistic patterns. This pretraining phase allows the transformer model to develop a deep understanding of language, making it capable of performing a wide range of downstream tasks, such as text classification. Table 2 showcases the performance of four transformer based models using word embeddings.

Table.2. Performance of four transformer based models using word embeddings.

| Transformer model | Accuracy | F1 score | Recall | Precision | AUC |
|---|---|---|---|---|---|
| BERT | 90% | 89.5% | 85% | 94.4% | 96.3% |
| XLNET | 90% | 89.9% | 80% | 100% | 97.7% |
| DistilBERT | 88.8% | 88.7% | 87.5% | 89.7% | 95.1% |
| RoBERTa | 91.3% | 90.6% | 85.0% | 97.1% | 97.1% |

These metrics provide insights into the performance of each transformer model. BERT achieved an accuracy of 90%, an F1 score of 89.5%, a recall of 85%, a precision of 94.4%, and an AUC of 96.3%. XLNET achieved an accuracy of 90%, an F1 score of 89.9%, a recall of 80%, a precision of 100%, and an AUC of 97.7%. DistilBERT achieved an accuracy of 88.8%, an F1 score of 88.7%, a recall of 87.5%, a precision of 89.7%, and an AUC of 95.1%. RoBERTa achieved the highest accuracy of 91.3%, an F1 score of 90.6%, a recall of 85.0%, a precision of 97.1%, and an AUC of 97.1%. These results highlight the performance of each transformer model in terms of accuracy, precision, recall, and AUC. RoBERTa demonstrated the highest accuracy and precision, while XLNET achieved a perfect precision score.

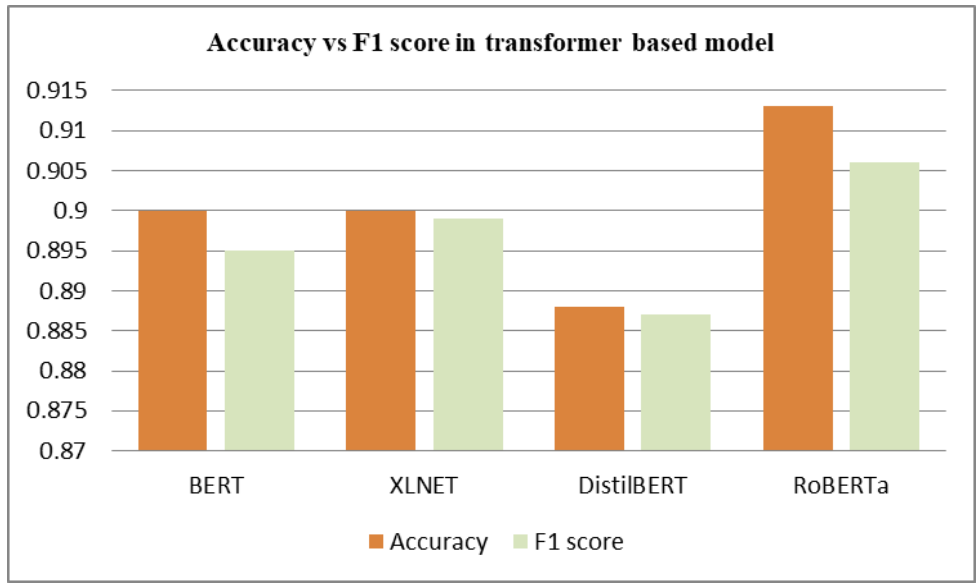

Figure 4. Comparison of all four transformer based model (accuracy vs F1 score) using barchart

Figure 4 presents a comprehensive overview of the transformer based model performance using word embeddings in a train-test split. The figure clearly illustrates that the RoBERTa outperforms all other transformers, achieving an impressive accuracy of 91.3%. Additionally, the bar chart shows that all four models have high accuracy, but there are some differences in F1 score. RoBERTa has the highest F1 score and accuracy, followed by BERT, XLNET, and DistilBERT.

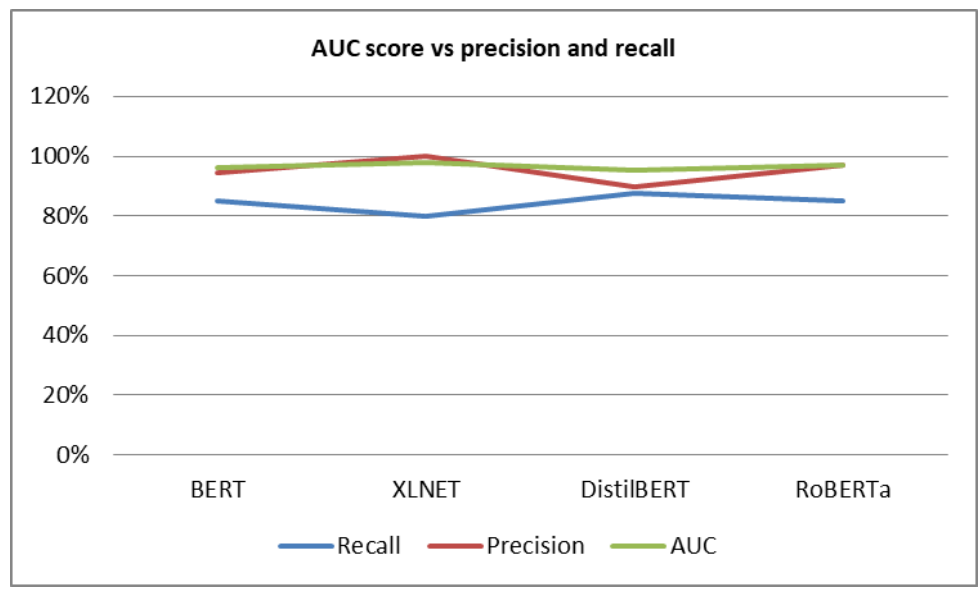

Figure 5. Comparison of all four transformer based model (AUC-ROC vs recall vs precision) using line graph

Figure 5 shows the chart of AUC- ROC curve vs precision vs recall. The graph shows that XLNET has the highest AUC score, followed by RoBERTa, DistilBERT, and BERT. RoBERTa has the highest precision and highest recall, followed by XLNET, BERT, and DistilBERT.

## 4.3  Comparison Analysis

Due to the limited size of the dataset, the machine learning model yielded satisfactory results. Various deep learning models were experimented with, including LSTM, GRU, and others, achieving a moderate accuracy of around 86%. As a result, we ultimately opted for pretrained transformer models such as BERT, RoBERTA, and similar ones. The transformer model exhibited superior performance in comparison. The performance comparison between machine learning models and transformer-based models can be found in figures 6 and 7.

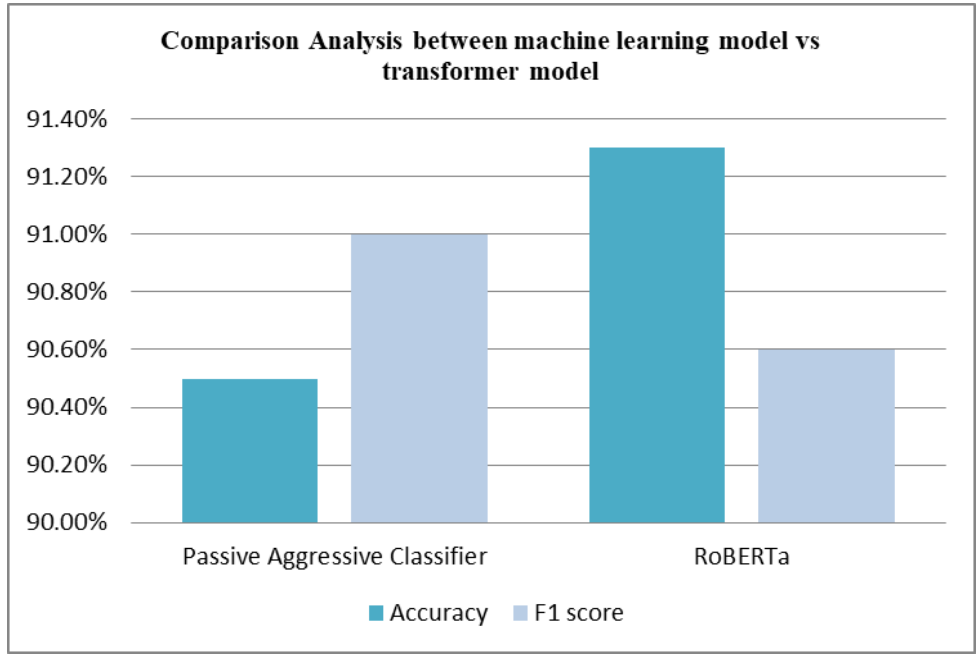

Figure 6. Comparison of  machine learning model vs transformer based model (accuracy vs F1 score)

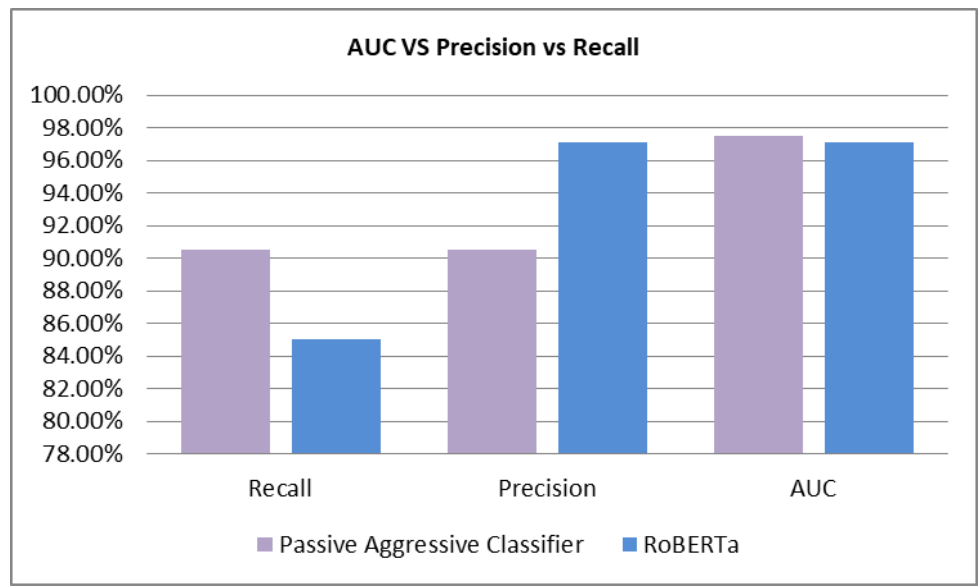

Figure 7. Comparison of machine learning model vs transformer based model (AUC vs Precision vs Recall)

## 5   Conclusion

Deceptive text has become a growing concern in recent years, particularly in the ecommerce industry and social media platforms, where fake reviews can mislead customers. Distinguishing between reliable and unreliable products based solely on reviews has become a challenge. To tackle this issue, our research paper focuses on detecting deceptive text using both machine learning and deep learning techniques. Initially, we explored various methods for feature extraction and word embeddings commonly employed by researchers in this field. We then delved into traditional machine learning approaches for deceptive text detection, providing summarized tables and charts showcasing their performance. Additionally, we conducted experiments using transformer-based models and compared their performance to the machine learning approach. Our findings revealed that the transformer-based model, specifically RoBERTa, achieved the highest accuracy on the deceptive text dataset. It outperformed the results obtained by the machine learning model, namely the passive aggressive classifier, by 0.8%. This demonstrates the effectiveness of utilizing transformer models for detecting deceptive text. By combining machine learning and deep learning techniques, we aim to enhance the detection and identification of deceptive text, enabling users to make more informed decisions and promoting a trustworthy online environment. The results of our comparative analysis highlight the potential of transformer-based models in addressing the challenges posed by deceptive text. In future, we have an idea to leverage collaborative filtering techniques to analyze user behavior, ratings, and interactions to detect anomalies or suspicious patterns that may indicate the presence of deceptive text.

## Acknowledgements

We thank the anonymous referees for their useful suggestions.